\documentclass{article}
\usepackage{ijcai25}

\usepackage{times}
\usepackage{soul}
\usepackage{url}
\usepackage[hidelinks]{hyperref}
\usepackage[utf8]{inputenc}
\usepackage[small]{caption}
\usepackage{graphicx}
\usepackage{amsmath}
\usepackage{amsthm}
\usepackage{amssymb}
\usepackage{booktabs}
\usepackage{algorithm}
\usepackage{algorithmic}
\usepackage[switch]{lineno}
\usepackage{color}
\usepackage{multirow}
\usepackage{dsfont} 

\newtheorem{property}{Property}

\title{Persistent Homology-induced Graph Ensembles for Time Series Regressions}

\author{
Viet T. Nguyen$^{1, 2}$ 
\and
Duy A. Pham$^{3, 4}$ 
An T. Le$^{5}$
Jans Peter$^{5, 6, 7}$\\ \And
Gunther Gust$^{1, 2}$
\affiliations
$^1$ University of Würzburg, Germany\\
$^2$ Center for Artificial Intelligence and Data Science (CAIDAS)\\
$^3$ Osnabrück University, Germany\\
$^4$ Leibniz Institute of Agricultural Engineering and Bio-economy (ATB)\\
$^5$ Technical University of Darmstadt, Germany\\
$^6$ German Research Center for AI (DFKI) \\
$^7$ Hessian.AI \\
}

\begin{document}

\maketitle

\begin{abstract}
The effectiveness of Spatio-temporal Graph Neural Networks (STGNNs) in time-series applications is often limited by their dependence on fixed, hand-crafted input graph structures. Motivated by insights from the Topological Data Analysis (TDA) paradigm, of which real-world data exhibits multi-scale patterns, we construct several graphs using \textit{Persistent Homology Filtration}---a mathematical framework describing the multiscale structural properties of data points. Then, we use the constructed graphs as an input to create an ensemble of Graph Neural Networks. The ensemble aggregates the signals from the individual learners via an attention-based routing mechanism, thus systematically encoding the inherent multiscale structures of data. Four different real-world experiments on seismic activity prediction and traffic forecasting (PEMS-BAY, METR-LA) demonstrate that our approach consistently outperforms single-graph baselines while providing interpretable insights. Our code implementation is provided in this \href{https://github.com/vietngth/ph-ensemble-gnn/}{URL}.
\end{abstract}

\section{Introduction and Related Work}
Many real-world applications benefit from learning from networked sensor data. Examples include, among others, traffic forecasting systems that are based on street sensors~\cite{jiang2022:trafficforeastingsurvey} or earthquake warning systems based on distributed seismic sensors~\cite{jozinovic2020:mle0,kim2021:mle1,bloemheuvel2023:mle_main}. Recent state-of-the-art approaches typically leverage graph neural networks~(GNNs)~{\cite{li2018:traffic_metrla_pemsbay_dcrnn,shao2022:D2STGNN,shao2022:STEP,jiang2023:MegaRCN,fan2024:rgdan,gao2024:STD-MAE,kim2021:mle1,bloemheuvel2023:mle_main}, due to their capability of modeling both temporal features from large-scale time series sensor data and spatial dependencies inherent in the sensor network. However, to apply these approaches, a key requirement is to generate an effective input graph.

One stream of works, subsumed commonly under the term graph structure learning, tries to overcome the graph generation problem by learning a graph representation jointly with a downstream task in an end-to-end deep learning pipeline \cite{zhang2020:gsl2,zhu2021:gsl1,GNNBook2022:gnnbook}. However, graph structure learning is computationally expensive given the large number of potential graphs. Therefore, another set of approaches tackles graph generation heuristically~\cite{GNNBook2022:gnnbook}. For example, a graph is generated by computing pairwise geographical distances of sensors, and a heuristic threshold is then applied to adjust the graph’s sparsity and control the distribution of its edges ~\cite{bloemheuvel2023:mle_main,li2018:traffic_metrla_pemsbay_dcrnn}. This threshold is either based on domain knowledge or chosen in an iterative manual process via trial-and-error and cannot fully capture the inherent complex dependencies of real-world sensor data. In summary, constructing an appropriate graph representation for the complex, potentially multi-scale patterns inherent in sensor network data is still an area open for research.   

In this work, we borrow from the theory of topological data analysis~\cite{chazal2021:tda_intro} to derive a novel ML architecture that is based on an ensemble of graph neural networks. Concretely, we use the concept of persistent homology~(PH), a data-driven mathematical framework that describes the evolution of topological features (e.g., connected components, holes, and voids) across different scales and resolutions \cite{zomorodian2004:ph_math1,edelsbrunner2008:ph_math2} (cf. Figure~\ref{fig:ph-example}a), to construct a task-specific set of effective input graphs. PH has been proven an effective tool for data analysis~\cite{de2007:ph_sensorhomological,li2019:ph_transporttopological,rieck2020:ph_medical_imaging2} and has been effectively used as a feature extractor to enhance learning representations~\cite{adams2017:persistence,townsend2020:ph_otherrepresentation,wang:2024ph_rep_persistent,ying:2024ph_rep_boosting}. Afterwards, we design ensembles of GNNs that encode these multiple graph inputs and aggregate their results for the downstream tasks. 

We test our architecture empirically on four data sets from two different real-world applications achieving state-of-the-art performance. First, we perform a time series extrinsic regression~(TSER) task~\cite{tan2021:tsertask} on two large-scale seismic datasets for early earthquake warning~\cite{michelini2016:seismic_dataset1,danecek2021:seismic_dataset2}. Second, we show competitive results on traffic forecasting tasks on two popular datasets METR-LA~\cite{jagadish2014:traffic_metrla_pre,li2018:traffic_metrla_pemsbay_dcrnn} and PEMS-BAY~\cite{li2018:traffic_metrla_pemsbay_dcrnn}. Finally, we leverage the advantageous architectural properties of our ensemble and analyze how  individual graphs contribute to the predictions, providing interesting insights into the application domains.
\section{Background}
\begin{figure*}[t]
    \centering
    \includegraphics[width=0.8\textwidth]{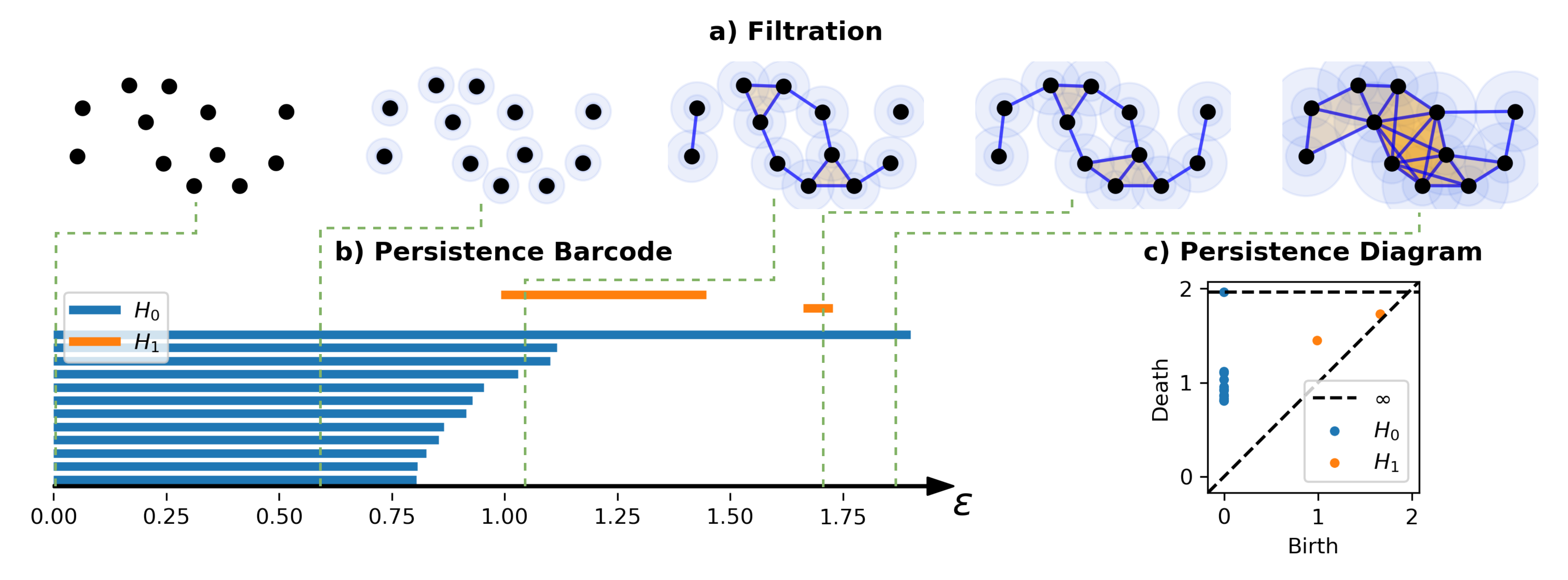}
    \caption{Computation step of persistent homology given data points. (a) A filtration that tracks the topological changes of the dataset with two homology classes $H_0$ (connected components) and $H_1$ (loops or tunnels). (b) The persistent barcode of the corresponding filtration. Epsilon ($\epsilon$) is the radius parameter that determines when points are connected - as it increases, points within $\epsilon$ distance of each other become connected, revealing the data's topological structure (c) The persistence diagram shows the lifespan of topological features---an equivalent representation of the barcode.}
    \label{fig:ph-example}
\end{figure*}
\begin{table}[tb]
    \centering
    \begin{tabular}{ccccc}
        \toprule
        Datasets  & \#Sensors & Sample Rate & Area (km$^2$)\\
        \midrule
        CI & 39 & 100 Hz & 714 \\
        CW & 39 & 100 Hz & 14835 \\
        METR-LA & 207 & 5 Min & 134 \\
        PEMS-BAY & 325 & 5 Min & 289 \\
        \bottomrule
    \end{tabular}
    \caption{Basic statistics of the datasets, including number of sensors, sample rate, and geographical coverage of the sensor networks.}
    \label{tab:dataset-description}
\end{table}
We first introduce notations of graph machine learning and provide a brief preliminary overview of persistent homology.
\subsection{Graph Construction of Sensor Network}\label{section3.1:graphprelim}
We consider sensor networks as undirected graphs. An undirected graph is defined as a pair $G=(V, E)$, where the vertex set $V$ and the edge set $E = \{ (u, v) | u,v \in V,\, u \neq v\}$ are finite $|V|=N, \,|E|=M$. Each edge is encoded with a weight $e_{ij} \in \mathbb{R}$. A distance matrix $\mathbf{\Gamma} \in \mathbb{R}^{N \times N}$ is denoted as $\mathbf{\Gamma} = \{m_{ij}|i, j\in V, m_{ij} = \text{Vincenty's distance}(i,j)\}$, which represents the pairwise geodesic distances of the vertices. Here, $m_{ij}$ is measured by Vincenty's distance based on Earth's ellipsoidal model~\cite{vincenty1975:geodesic}. The edge weights are normalized by the smallest and largest distances $m_{\min}, m_{\max} \in \mathbf{\Gamma}$, i.e.,
 \begin{equation}
     e_{ij} = 1 - \frac{e_{ij} - m_{\min}}{m_{\max} - m_{\min}}.
 \end{equation}
This assumes that sensors in closer proximity exchange information more rapidly and effectively within the network, thus resulting in higher weight assignments. Then, we define the weighted adjacency matrix $\mathbf{A} \in \mathbb{R}^{N \times N}$ as:
\begin{equation}
    {\*\mathbf{A}}_{ij} = \begin{cases}
    e_{ij}, & \text{if} \; e_{ij} < \tau \\
    0, & \text{otherwise,}
\end{cases}
\end{equation}
where $\tau \in (0, 1)$ is a given threshold, and $e_{ij} > 0$ indicates the presence of an edge between vertices $i$ and $j$. The neighborhood of a vertex $v \in V$ is defined as $N(v)=\{u \in V | (v, u) \in E \}$, and the degree of $v$ is $\text{deg}(v) = \sum_{u \in N(v)} \mathds{1}_{\mathbf{A}_{vu} > 0}$, where $\mathds{1}_{\mathbf{A}_{vu} > 0}$ is the indicator function that equals $1$ if $\mathbf{A}_{vu} > 0$, and $0$ otherwise. Finally, we define the diagonal degree matrix $\mathbf{D} \in \mathbb{R}^{N \times N}$, whose entry $\mathbf{D}_{ii}$ corresponds to the degree of vertex $i$, i.e., $\mathbf{D}_{ii} = \sum^N_{j=i} \mathds{1}_{\mathbf{A}_{ij} > 0}$. 
\subsection{Graph Convolutional Neural Networks}
GNNs are designed to encode relational data such as graphs into a lower-dimensional embedding space, where each node is represented by a real-valued vector. These node representations are iteratively updated by exchanging and aggregating local information using the message-passing mechanism within the network topology~\cite{scarselli2008:original_gnn,gilmer2017:neuralmsp}. In this study, we utilize a simple, yet effective graph convolutional network (GCN)~\cite{Kipf2017:gcn} as the graph encoder. Let $\mathbf{X} \in \mathbb{R}^{N \times N}$ be the input feature matrix, where each node is annotated with a $d-$dimensional vector that encodes initial information such as temporal features from time series. The message-passing function of the $(l+1)-$layer $(l>0)$ of a GCN is defined as:
\begin{equation} \label{eq:gcn}
    \mathbf{H}_{l+1} = \mathbf{\sigma}(\mathbf{\Tilde{D}}^{-\frac{1}{2}}\mathbf{\Tilde{A}} \mathbf{\Tilde{D}^{-\frac{1}{2}}} \mathbf{H}_l\mathbf{W}_l),
\end{equation}
where $\mathbf{H}_0  = \mathbf{X}$, and $\Tilde{\mathbf{A}} = \mathbf{A} + \mathbf{I}_N$ is the adjacency matrix with added self-loops, which allows each node to leverage its own features for updating the node representations. Here, $\mathbf{I}_N \in \mathbb{R}^{N \times N}$ is the identity matrix, $\mathbf{\Tilde{D}}$ is a diagonal matrix with $\mathbf{\Tilde{D}}_{ii} = \sum_j \mathbf{\Tilde{A}}_{ij}$, and $\mathbf{W}_l \in \mathbb{R}^{F \times F'}$ is a learnable linear transformation matrix, where $F$ and $F'$ denote the embedding dimensions in layers $l$ and $(l+1)$, respectively. The output $\mathbf{H}_l$ is passed through an element-wise non-linear activation function $\mathbf{\sigma(.)}$. The final representations $\mathbf{H}_L$ in the last layer are used for downstream tasks such as node regressions.
\subsection{Basic Concepts in Topological Data Analysis}
In this section, we present an overview of relevant concepts in TDA. Readers can refer to~\cite{zomorodian2004:computing_persistent_homology,hatcher2002:algebraic} for more formal and rigorous definitions.
\paragraph{Simplicial Complex.} Extracting topological information from a set of data points is inherently challenging. To address this, a $k-$simplicial complex is constructed as a proxy for the underlying shape of the sampled points. This simplicial complex serves as a higher-dimensional extension of a graph, which comprises a collection of simplicies of varying dimensions. The $k-$simplicial complex has geometric realizations include vertices $(k=0)$, edges $(k=1)$, filled triangular faces $(k=2)$, solid tetrahedra $(k=3)$, and analogous higher-dimensional shapes $(k \geq 4)$.
\paragraph{Homology.}Homology is a group in algebraic topology that quantifies the topological features of the simplicial complexes across multiple dimensions, i.e., structures like ``holes". The $d-$dimensional holes of a topological space $X$ are captured by its $d-$th homology group, denoted as $H_d(X)$. These ``holes" can exist in various dimensions: connected components $(H_0)$, loops or tunnels $(H_1)$, and enclosed voids $(H_{d\geq 2})$. The size of this group, measured by its rank, is known as the $d-$th Betti number, $\beta_d$, which represents the number of independent $d-$dimensional features in $X$. 
\paragraph{Vietoris-Rips Complex.} Simplicial complexes can be constructed by the Vietoris-Rips (VR) method that approximates the topology of a dataset with a designated radius~\cite{gromov1987:hyperbolic,hatcher2002:algebraic,munkres2018:elements}. Given a set of points $X \subset \mathbb{R}^n$ and a fixed radius $\epsilon > 0$, we start by considering a sphere of radius $\epsilon$ around each point in $X$. A collection of points forms a higher-dimensional geometric object if all the points in that group are pairwise connected by distances no greater than $\epsilon$. To construct the VR complex, we use the distance matrix $\mathbf{\Gamma}$ defined in Section~\ref{section3.1:graphprelim}. A $d-$dimensional object is included if:
\begin{equation}
 \max_{1\leq i, j \leq k+1} \mathbf{\Gamma}_{ij} < \epsilon.   
\end{equation}
In Figure~\ref{fig:ph-example}a, the middle figure depicts a VR complex with 3 connected components $(H_0)$ and 1 tunnel $(H_1)$. The corresponding Betti numbers are $\beta_0 = 3$ and $\beta_1 = 1$.
\subsection{Persistent Homology}
Persistent homology (PH) examines how topological features---such as connected components, loops, and voids---emerge and disappear across scales in a \textit{filtration}, which is a sequence of nested simplicial complexes built by gradually increasing the scale parameter $\epsilon$. Features are said to be ``born" when they appear and ``die" when they vanish, thus providing insights into the multi-scale structure of the data.

Given a filtration $C=(C_i)_{i\geq 0}$ of simplicial complexes and their homology groups $H_d(C_i)$, PH computes the $d-$th persistent homology groups, defined as the images of the induced maps between homology groups:
\begin{equation}
    H_p^{i \to j}(C) = \text{Im} \big( H_p(C_i) \to H_p(C_j) \big), 0 \leq i \leq j, \ p \geq 0.
\end{equation}
The persistent Betti numbers $\beta_d^{i, i}$ are the ranks of these groups and quantify the number of $d-$dimensional features that persist from $C_i$ to $C_j$. An example of filtration is provided in Figure~\ref{fig:ph-example}a. Typically, $d$ is set to $d:=2$ due to computational efficiency (connected components and tunnels). 

The lifespans of the features are tracked using barcodes (Figure~\ref{fig:ph-example}b) or persistent diagrams (PDs) (Figure~\ref{fig:ph-example}c), where births are plotted on the $x-$axis and deaths on the $y-$axis. Features with longer lifespans that appear as longer bars or farther from the diagonal represent more significant features. 

\section{Methodology} \label{sec:method}

We propose a multi-graph construction method using PH filtration and design ensembles of neural network representations for two tasks: earthquake predictions and traffic forecasting, where sensor networks are represented as geometric graphs based on their topology.
\begin{algorithm}[tb]
    \caption{Graph Generation with Persistent Homology}
    \label{alg:graph-generation}
    \textbf{Input}: Distance matrix $D$, Sensor set $S$\\
    \textbf{Output}: PH-induced graphs
    \begin{algorithmic}[1]
        \STATE Extract persistent diagram $PD$ by computing PH of $D$ using Vietoris-Rips complex
        \STATE Initialize $E^{0,1} \gets$ death times in $PD$; $G^0, G^1, G^{0,1} \gets \emptyset$
        \FOR{each $\epsilon_t \in E^{0,1}$}
            \STATE $G^t \gets (S, \{(s_i, s_j) \mid D_{ij} \leq \epsilon_t\})$
            \IF{$\epsilon_t$ is a 0-Dim homology death time}
                \STATE $G^0 \gets G^0 \cup G^t$
            \ELSE
                \STATE $G^1 \gets G^1 \cup G^t$
            \ENDIF
            \STATE $G^{0,1} \gets G^{0,1} \cup G^t$
        \ENDFOR
    \RETURN $G^0, G^1, G^{0,1}$
    \end{algorithmic}
\end{algorithm}
\subsection{Graph Generation with Persistent Homology}
\begin{figure}[ht]
    \centering
    \includegraphics[width=0.8\columnwidth]{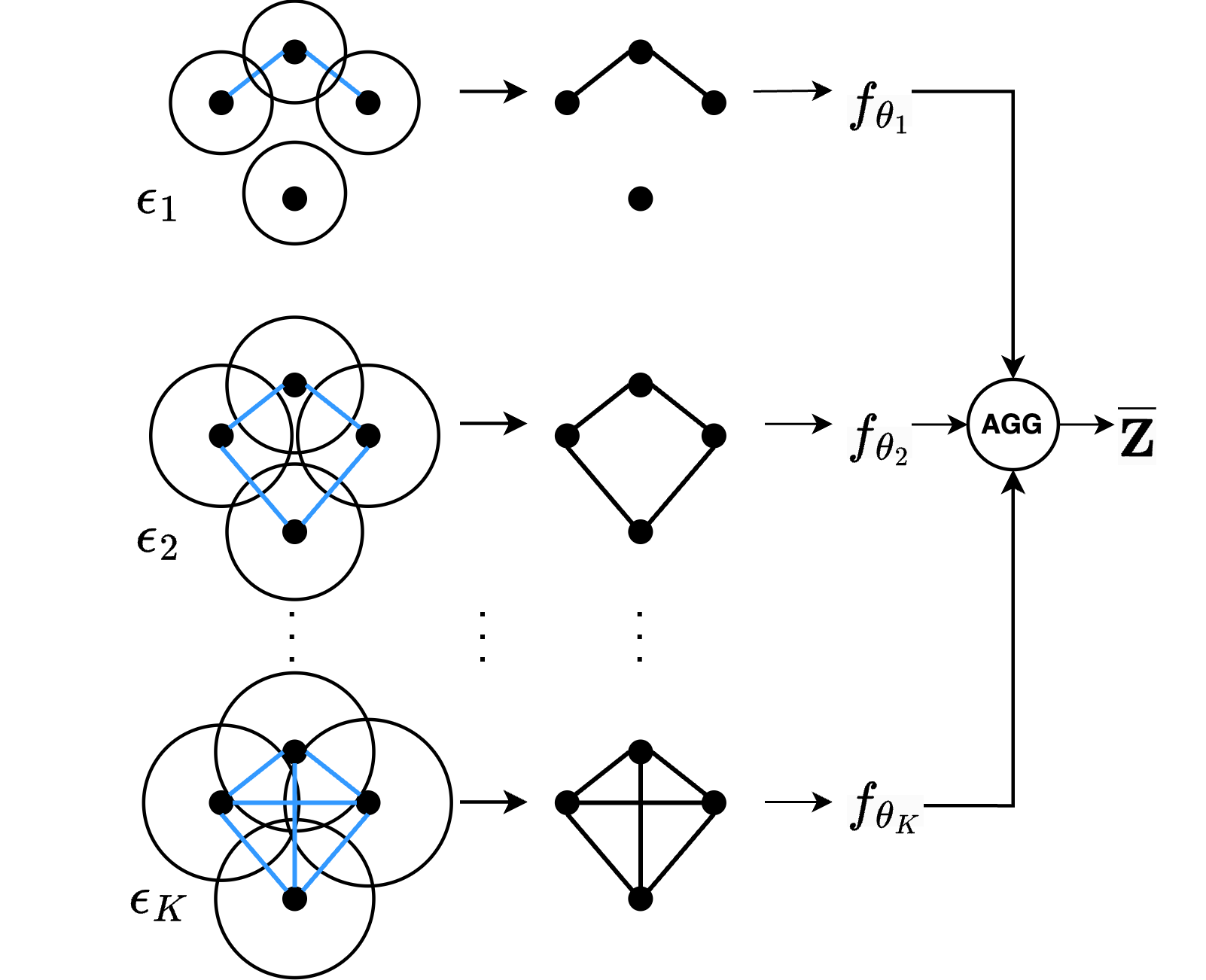}
    \caption{Graph generation using persistent homology. Each graph is generated by the filtration at the threshold $\epsilon_t$. The graphs are fed separately into arbitrary functions $f_{\theta_k}$ parametrized by learnable weights $\theta_k$ (e.g., neural networks), which are then aggregated ($AGG$) to obtain the representations $\mathbf{\overline{Z}}$ for the downstream tasks.}
    \label{fig:phgnn}
\end{figure}
The ``death times" in the filtration mark critical thresholds where the data's structure undergoes topological changes. To leverage this, we construct a multi-scale representation of the data as a series of discrete graphs, where each captures the topology at a specific scale defined by its death times. To integrate this representation with GNNs, we extract the ``skeleton" of the simplicial complexes by considering only pairwise node connections. Algorithm~\ref{alg:graph-generation} details this process. We generate three graph sets: $G^0$ having graphs whose simplicial complexes contain only $H_0$; $G^1$ having graphs whose simplicial complexes contain both $H_0$ and $H_1$; and $G^{0,1}$ being the union of $G^0$ and $G^1$. For instance, in Figure~\ref{fig:phgnn}, graphs at $\epsilon_1$ and $\epsilon_2$ belong to $G^0$, while the graph at $\epsilon_K$ is part of $G^1$.

More formally, given a dataset \( X = \{x_1, \ldots, x_n\} \subseteq \mathbb{R}^m \) and its associated persistent homology, we construct graphs based on the Vietoris-Rips filtration \(C = \{C_t\}_{t \geq 0}\). For each \(d\)-dimensional topological feature, such as connected components for \(d=0\) or loops for \(d=1\), we identify the set of death times \(\{\tau_i^d\}_{i=1}^{b_d}\), where \( b_d \) denotes the total number of these features in dimension \(d\). At each death time \(\tau_i^d\), we define a graph \(G_i^d = (V, E_i^d)\), with \(V = X\) as the vertex set and \(E_i^d\) containing edges between points \(x_\alpha\) and \(x_\beta\) if their pairwise distance satisfies \(\|x_\alpha - x_\beta\| \leq \tau_i^d\). The resulting collection of graphs, \(\mathcal{G}^d = \{G_i^d\}_{i=1}^{b_d}\), represents the data’s topological features across multiple scales, capturing relationships from fine-grained local structures to global patterns. 

\subsection{Model Architecture: Ensembles of GNNs}
We design ensembles of neural networks to encode the PH-induced graphs, where each sub-network takes a distinct graph input. The general architecture of the $i-$th sub-network processing the $i-$th graph with the respective adjacency matrix $\mathbf{A}^{(i)}$ is defined as:
\begin{align} 
    \mathbf{X}^{(i)} &= \sigma_1 \left(f^{(i)}\left(\mathbf{D};  \mathbf{\Theta}^{(i)}\right)\right),\label{eq:en1}\\ 
    \mathbf{H}^{(i)} &= \sigma_2\left(g^{(i)}\left(\mathbf{X}^{(i)}, \mathbf{A}^{(i)}; \mathbf{\Psi}^{(i)}\right)\right), \label{eq:en2}\\ 
    \mathbf{\overline{Z}} &= \bigoplus_i \mathbf{Z}^{(i)}\label{eq:en3}, \\
    \mathbf{\overline{Y}} &= \text{MLP}(\mathbf{\overline{Z}};\mathbf{\Phi}) \label{eq:en4},
\end{align}
where $\sigma_1$ and $\sigma_2$ are non-linear activation functions, and $f_i(.)$ is an arbitrary function chosen depending on the applications to extract the temporal features from the time series. The function $g_i(.)$ is the 2-layer GCN model where each message-passing layer is defined as in Eq~\ref{eq:gcn}. The encodings $\mathbf{H}^{(i)}$ are set to be the node representations of the GCN's last layer. The matrices $\mathbf{\Theta}^{(i)}, \mathbf{\Psi}^{(i)}$ are learnable weights of $f^{(i)}$ and $g^{(i)}$, and $\mathbf{\Phi}$ is learnable weights shared across subnetworks. The operator $\bigoplus$ defines an aggregator to fuse the output to obtain the final representation for the downstream task. Here, we adapt the attention-based mechanism~\cite{bahdanau2015:att} to guide the model to select the most important graphs using the outputs of the GCNs as weights:
\begin{align}
    \mathbf{S}^{(i)} &= \mathbf{H}^{(i)}\mathbf{W}_{att},  \label{eq:attn1}\\
    \boldsymbol{\alpha}^{(i)} &= \frac{\exp(\mathbf{S}^{(i)})}{\sum_j \exp(\mathbf{S}^{(j)})}, \label{eq:attn2}\\
    \mathbf{\overline{Z}} &= \sum_i \boldsymbol{\alpha}^{(i)}  \odot \mathbf{Z}^{(i)}. \label{eq:attn3},  
\end{align}
where $\mathbf{W}_{att}$ is trainable weights shared across sub-networks, and $\odot$ denotes the Hadamard product. The representations are passed through a multilayer perceptron (MLP) to adapt to output dimensions for the node-level regression tasks. 
\subsection{Theoretical Properties} \label{subsec:theory}
In order to analyze GNN ensemble representation guided by PH, we begin by defining the core topological structures. For data points $\mathbf{X}$ and target labels $\mathbf{Y}$ sampled from manifold $\mathcal{M}$, we define the topological signal at scale $t$ as:
\begin{equation}
    S_t(\mathbf{X}) = \{\sigma \in H_*(\mathbf{X}_t) \mid \sigma \text{ is a homology class at scale } t\}
\end{equation}
where $\mathbf{X}_t$ represents the Vietoris-Rips complex at scale $t$.


For simplicity, we consider an arbitrary GNN architecture. For an ensemble of graphs derived from either PH or arbitrary thresholds, we construct an ensemble representation by concatenating their respective GNN-derived features. Specifically, let $\mathbf{F}_t \in \mathbb{R}^d$ be the GNN feature at threshold $t$, then $\mathbf{F}_{\text{PH}} = [\mathbf{F}_{\tau^d_1} \parallel \cdots \parallel \mathbf{F}_{\tau^d_m}]$ and $\mathbf{F}_{\text{arb}} = [\mathbf{F}_{t_1} \parallel \cdots \parallel \mathbf{F}_{t_m}]$ using death times $\{\tau^d_i\}_{i=1}^m$ and arbitrary thresholds $\{t_i\}_{i=1}^m$ respectively. The efficacy of these representations can be quantified through the mutual information between the learned features and target labels $\mathbf{Y}$:

\begin{equation}
  I(\mathbf{F};\mathbf{Y}) = \mathbb{E}_{p(\mathbf{f},\mathbf{y})}\left[\log\frac{p(\mathbf{f},\mathbf{y})}{p(\mathbf{f})p(\mathbf{y})}\right],
\end{equation}
where $p(\mathbf{f},\mathbf{y})$ denotes the joint probability density and $p(\mathbf{f}), p(\mathbf{y})$ are the corresponding marginal densities. 

\subsubsection{Information Preservation of PH-induced Graphs.}
Then, we present that a key property of our persistent homology framework is its optimal information preservation, underlined by the following property:
\begin{property}
    PH-induced thresholds maximize information preservation.
\end{property}
Here, we assume that the input data follows an isotropic Gaussian distribution $\mathcal{N}(\boldsymbol{\mu}, \sigma^2\mathbf{I}_m)$ centered on a compact Riemannian manifold $\mathcal{M}$, with labels generated by a smooth function $f:\mathcal{M} \rightarrow \mathcal{Y}$ and persistent homology computations employing standard Euclidean distance metrics. To motivate the use of PH for graph ensemble construction, we first examine when topological information is most significant. At death time $\tau^d_i$ in the filtration, a homology class disappears, creating a change in topological signal $\Delta S_{\tau^d_i} = S_{\tau^d_i + \epsilon} - S_{\tau^d_i - \epsilon}$. For a set $\mathbf{X}$ sampled from manifold $\mathcal{M}$ with labels $\mathbf{Y}$ dependent on $\mathcal{M}$'s topology, the properties of Vietoris-Rips filtration ensure that homology groups remain constant between death times: $H_*(\mathbf{X}_{t_1}) \cong H_*(\mathbf{X}_{t_2})$ for any $t_1,t_2$ between consecutive death times. Thus, while $\Delta S_t = 0$ for $t \neq \tau^d_i$, changes at death times capture precise topological transitions, leading to:
\begin{equation}
    I(\Delta S_{\tau^d_i}; \mathbf{Y}) > I(\Delta S_t; \mathbf{Y}) \quad \text{for } t \neq \tau^d_i \label{eq:IDeltaTau}
\end{equation}
This motivates our choice of death times as filtration values for constructing graph sequences, as each captures a distinct, information-rich topological transition in the data structure.

\subsubsection{Information Preservation of Ensembling.}
Building on the informativeness of PH-derived thresholds, we consider how to optimally combine the resulting graph representations based on the following property.
\begin{property}
    Instance-dependent feature weighting achieves optimal information preservation.
\end{property}
The mutual information of the PH-induced graph ensemble feature and the labels can be factorized due to concatenation
\begin{equation}
    I(\mathbf{F}_{\text{PH}}; \mathbf{Y}) = \sum_{i} I(\mathbf{F}_i; \mathbf{Y}) - \sum_{i,j} I(\mathbf{F}_i; \mathbf{F}_j; \mathbf{Y}),
\end{equation}
which is derived from the chain rule of mutual information and the nested property of Vietoris-Rips filtration. This leads to features derived from distinct death times exhibiting approximate conditional independence given the labels:
\begin{equation}
    I(\mathbf{F}_i; \mathbf{F}_j | \mathbf{Y}) < \epsilon
\end{equation}
for some small $\epsilon > 0$.

Then, we show the information capacity advantages of PH-induced ensembles. First, we show that instance-dependent feature weighting achieves optimal information preservation.  This is formalized through a loss function that balances information capture with feature redundancy:
\begin{equation}
    \mathcal{L}(\mathbf{w}) = -\underbrace{I\left(\sum_i w_i(\mathbf{x})\mathbf{F}_i; \mathbf{Y}\right)}_{\text{information capture}} + \lambda \underbrace{\sum_{i,j} w_i(\mathbf{x})w_j(\mathbf{x})I(\mathbf{F}_i; \mathbf{F}_j)}_{\text{redundancy penalty}}
\end{equation}
The first term maximizes feature-label mutual information, while the second penalizes redundancy between features. Such instance-dependent weighting performs at least comparable to any static weighting scheme. By exploiting the convexity of negative mutual information and positive semi-definiteness of the feature interaction term, optimizing $\mathcal{L}(\mathbf{w})$ guarantees:
\begin{equation}
    I\left(\sum_i w_i(\mathbf{x})\mathbf{F}_i; \mathbf{Y}\right) \geq I\left(\sum_i c_i \mathbf{F}_i; \mathbf{Y}\right)
\end{equation}
for any static weights $\{c_i\}$.

Second, we establish that PH-induced thresholds strictly outperform arbitrary thresholds in information preservation. By leveraging the properties of persistent homology filtration, the change in topological signal $\Delta S_t$ is maximized at death times $\tau^d_i$, motivated directly by Eq. \ref{eq:IDeltaTau} leading to
\begin{equation}
    I(\mathbf{F}_{\text{PH}}; \mathbf{Y}) \geq I(\mathbf{F}_{\text{arb}}; \mathbf{Y}) + \Delta,
\end{equation}
where $\Delta > 0$ quantifies the information gained from capturing true topological transitions rather than arbitrary structural changes. The mutual information gain of $\mathbf{F}_{\text{PH}}$ arises because death times mark significant, non-redundant changes in the homology of the underlying data manifold.
\subsection{Earthquake Predictions}
Time-series Extrinsic Regression (TSER) tasks study the relationship between the entire time series sequence and external variables~\cite{tan2021:tsertask}. The objective of earthquake predictions is to regress five seismic variables, known as \textit{maximum intensity measurements} (IMs) for each seismic sensor given their multivariate time series input. 

Recent work~\cite{bloemheuvel2023:mle_main} has shown that GNNs have proven effective by enabling sensors to exchange information, improving the accuracy of IM predictions even for distant sensors. To demonstrate the effectiveness of PH-induced graphs, we keep the architecture design's effort minimal, with each sub-network in the ensemble following a similar architecture as in their work. In particular, $f^{(i)}$ is a 2-layer 1D convolutional layers~\cite{kiranyaz2021:1dcnn} to extract the temporal features, with $\sigma_1(.)$ as the ReLU function. The output of the GCN's first layer passes through a ReLU function, while the final layer's output is processed using a Tanh function, i.e., $\text{Tanh}(x)=\frac{\exp^x - \exp^{-x}}{\exp^x + \exp^{-x}}$. The output is then fed into 5 separate MLPs, where each predicts one IM value.

The time series tensor is defined as $\mathbf{D} \in \mathbb{R}^{T \times N \times W \times C}$, where $T$ denotes the number of earthquakes, $W$ is the input length of the time series, $N$ is the number of sensors, and $C$ is the number of channels. The objective is to minimize a mean-square error (MSE) loss:
\begin{equation}
    \mathcal{L}_{tser} = \frac{1}{N} \| \overline{\mathbf{Y}} - \mathbf{Y} \|_2^2 + \lambda \| \mathbf{W} \|_2^2 
\end{equation}
where $\mathbf{Y} \in \mathbb{R}^{N \times 5}$ are the labels, and the second term is an L2 regularizer, with $\mathbf{W}$ being the entire model's weights and $\lambda=10^{-4}$ controlling the regularization strength.
\subsection{Traffic Forecasting}\label{sec:traffic-forecasting}
We study the traffic speed forecasting task using historical data of a network of traffic sensors, represented as nodes in a graph $G$. The traffic data is structured as a multivariate time series tensor $\mathbf{D} \in \mathbb{R}^{T_{\text{in}} \times N \times K}$, where $T_{\text{in}}$ is the number of historical time steps, $N$ is the number of sensors, and $K=1$ is the traffic speed at each sensor. The aim is to develop a mapping function $f$ that predicts the next $T_{\text{out}}$ time steps of traffic conditions:
\begin{equation}
    f(\mathbf{D}_{t-T_{\text{in}}+1:t}, G) = \mathbf{D}_{t+1:t+T_{\text{out}}},
\end{equation}
where $\mathbf{D}_{t+1:t+T_{\text{out}}}$ represents the predicted traffic conditions for $T_{\text{out}}$ steps ahead. The function $f^{(i)}$ in Eq~\ref{eq:en1} is modeled using a linear transformation followed by a one-layer Gated Recurrent Unit (GRU)~\cite{cho2014:gru} to capture sequential patterns of traffic time series. To model spatial dependencies across the network, the same GCN as in the Earthquake predition task is used. The objective of the model is to minimize the L1 loss:
\begin{equation}
\mathcal{L}_{traffic} = \frac{1}{N} \sum_{i=1}^N \| \overline{\mathbf{Y}} - \mathbf{Y} \|_1
\end{equation}
where $\mathbf{Y} \in \mathbb{R}^{T_{\text{out}} \times N \times 1}$ is the ground truth of traffic signals.
\begin{table*}
    \centering
    \begin{tabular}{lcccccccc}
        \toprule
        \multirow{2.5}{*}{Model} & \multirow{2.5}{*}{Graph Order(s)} & \multirow{2.5}{*}{\#Graphs (CI, CW)} & \multicolumn{3}{c}{Central Italy (CI)} & \multicolumn{3}{c}{Central-West Italy (CW)}  \\
        \cmidrule(lr){4-6} \cmidrule(lr){7-9} & & 
        & MAE& MSE & RMSE & MAE & MSE & RMSE \\
        \midrule \midrule
        JOZ-CNN& - & $(1, 1)$ & $0.35$ & $0.25$ & $0.48$ & $0.45$ & $0.37$ & $0.59$ \\
        KIM-GNN& - & $(1, 1)$ & $0.29$ & $0.24$ & $0.48$ & $0.46$ & $0.37$ & $0.6$ \\ 
        TSER-GCN& - & $(1, 1)$ &  $0.31$ & $0.20 $& $0.44$ & $0.41$ & $0.31$ & $0.54 $\\ 
        \midrule 
        PH-TSER-Max$_0$ & $0$ & $(38, 38)$ & $\textbf{0.28}$ & $\textbf{0.16}$ & $0.40$ & $0.35$ & $0.20$ & $0.45$ \\
        PH-TSER-Max$_{1}$ & $1$ & $(6, 8)$ & $0.29$ & $0.18$ & $0.42$ & $0.33$ & $0.18$ & $0.42$ \\
        PH-TSER-Max$_{0,1}$ & $0, 1$ & $(44, 46)$ & $\textbf{0.28}$ & $\textbf{0.16}$ & $0.40$ & $0.32$ & $0.17$ & $0.41$ \\
        \midrule
        PH-TSER-Mean$_{0}$ & $0$ & $(38, 38)$ & $0.29$ & $0.17$ & $0.41$ & $0.34$ & $0.20$ & $0.44$ \\
        PH-TSER-Mean$_{1}$  & $1$ & $(6, 8)$ & $0.29$ & $\textbf{0.16}$ & $0.40$ & $0.32$ & $0.16$ & $0.41$ \\
        PH-TSER-Mean$_{0,1}$ & $0, 1$ & $(44, 46)$ & $\underline{\textbf{0.27}}$ & $\underline{\textbf{0.15}}$ & $\underline{\textbf{0.39}}$ & $0.33$ & $0.18$ & $0.43$ \\
        \midrule
        PH-TSER-Att$_{0}$ & $0$ & $(38, 38)$ & $\textbf{0.28}$ & $\textbf{0.16}$ & $\underline{\textbf{0.39}}$ & $\underline{\textbf{0.30}}$ & $\underline{\textbf{0.15}}$ & $\underline{\textbf{0.39}}$ \\
        PH-TSER-Att$_{1}$ & $1$ & $(6, 8)$ & $0.29$ & $\textbf{0.16}$ & $0.40$ & $0.35$ & $0.20$ & $0.45$ \\
        PH-TSER-Att$_{0,1}$ & $0, 1$ & $(44, 46)$ & $\underline{\textbf{0.27}}$ & $\underline{\textbf{0.15}}$ & $\underline{\textbf{0.39}}$ & $\textbf{0.31}$ & $\textbf{0.16}$ & $\textbf{0.40}$ \\
        \bottomrule
    \end{tabular}
    \caption{Average of MAE, MSE, and RMSE of 5 metrics of the proposed models. Results of individual metrics are reported in the Appendix.}
    \label{tab:tser}
\end{table*}
\begin{table*}[ht]
    \centering
    \begin{tabular}{lccc|ccc}
        \toprule
        \multirow{2.5}{*}{Model} & \multicolumn{3}{c|}{METR-LA ($15/30/60$ min)} & \multicolumn{3}{c}{PEMS-BAY ($15/30/60$ min)}  \\
        \cmidrule(lr){2-4} \cmidrule(lr){5-7} 
        & MAE & MAPE(\%) & RMSE & MAE & MAPE(\%)& RMSE  \\
        \midrule \midrule
        DCRNN& $2.77/3.15/3.60$ & $7.30/8.80/10.50$ & $5.38/6.45/7.59$ & $1.38/1.74/2.07$ & $2.90/3.90/4.90$ & $2.95/3.97/4.74$\\
        D2STGNN & $2.56/2.90/3.35$ & $6.48/\textbf{7.78}/\textbf{\underline{9.40}}$ & $\textbf{4.88}/5.89/7.03$ & $1.24/1.55/1.85$ & $2.58/3.49/4.37$ & $\textbf{2.60}/3.52/4.30$\\
        MegaRCN & $2.52/2.93/3.38$ & $\textbf{6.44}/7.96/9.72$ & $4.94/6.06/7.23$ & $1.28/1.60/1.88$ & $2.67/3.57/4.41$ & $2.72/3.68/4.42$\\
        STEP & $2.61/2.96/3.37$ & $6.60/7.96/9.61$ & $4.98/5.97/\textbf{6.99}$ & $1.26/1.55/\textbf{1.79}$ & $2.59/3.43/\textbf{4.18}$ & $2.73/3.58/\textbf{\underline{4.20}}$\\
        STD-MAE & $2.62/2.99/3.40$ & $6.70/8.04/9.59$ & $5.02/6.07/7.07$ & $\textbf{1.23}/\textbf{1.53}/\textbf{\underline{1.77}}$ & $2.56/3.42/\textbf{\underline{4.17}}$ & $2.62/3.53/\textbf{\underline{4.20}}$ \\
        RGDAN & $2.69/2.96/3.36$ & $7.14/8.07/\textbf{9.54}$ & $5.20/5.98/7.02$ & $1.31/1.56/1.82$ & $2.77/3.47/4.28$ & $2.79/3.55/\textbf{\underline{4.20}}$ \\ 
        \midrule
        PH-FC-Att$_0$ & $\textbf{\underline{2.47}}/\textbf{2.83}/\textbf{3.40}$ & $\textbf{\underline{6.31}}/\textbf{\underline{7.57}}/9.71$ & $4.89/\textbf{5.79}/7.03$ & $\textbf{\underline{1.20}}/\textbf{\underline{1.51}}/1.94$ & $\textbf{\underline{2.42}}/\textbf{\underline{3.20}}/4.42$ & $\textbf{\underline{2.47}}/\textbf{3.28}/4.36$\\
        PH-FC-Att$_1$ & $3.19/3.49/3.99$ & $9.91/10.94/12.69$ & $5.40/6.13/7.20$ & $1.36/1.63/2.01$ & $2.91/3.59/4.68$ & $2.63/3.34/4.31$ \\
        PH-FC-Att$_{0,1}$ & $\textbf{2.50}/\textbf{\underline{2.82}}/\textbf{\underline{3.34}}$ & $6.74/7.89/9.76$ & $\textbf{\underline{4.77}}/\textbf{\underline{5.60}}/\textbf{\underline{6.75}}$ & $\textbf{1.23}/\textbf{\underline{1.51}}/1.90$ & $\textbf{2.52}/\textbf{3.24}/4.37$ & $\textbf{\underline{2.47}}/\textbf{\underline{3.24}}/\textbf{4.24}$ \\
        \bottomrule
    \end{tabular}
    \caption{Performance on METR-LA and PEMS-BAY. The results of the baselines are taken directly from the original papers.}
    \label{tab:traffic-forecasting}
\end{table*}
\section{Experiments}
In this section, we describe the experimental setups. 
The overall description of the datasets is provided in Table~\ref{tab:dataset-description}. 
\subsection{Earthquake Predictions}
We test on two earthquake datasets~\cite{michelini2016:seismic_dataset1,danecek2021:seismic_dataset2}. 
\paragraph{Datasets.} The data comprises continuous seismic wave amplitude records across three ground-motion channels. Waveforms are initially captured at epicenters, with time delays observed in sensors farther away. Key IMs include PGA for peak ground shaking, PGV for structural damage via seismic energy, and SA at periods (0.3s, 1s, 3s) for structural response. The Central Italy (CI) Network~\cite{danecek2021:seismic_dataset2} includes $N=39$ stations and $T=915$ earthquakes, while the Central-West Italy (CW) Network~\cite{michelini2016:seismic_dataset1} has $N=39$ stations and $T=266$ records, both with $C=3$ channels. Each station logs $W=10s$ of data for earthquakes from 01/01/2016 to 29/11/2016. CW poses greater challenges due to its wider, sparser sensor distribution. Further details are described in~\cite{jozinovic2020:mle0}.
\paragraph{Models.}Our variants were annotated as PH-TSER-AGGR$_k$, where AGGR denotes different aggregators $\bigoplus$ in Eq~\ref{eq:en4}. We compared basic operations such as Max, and Mean against the attention-based (Att) operator defined in Eq~\ref{eq:attn3}, and $k$ indicates graph set $G^{k}$ used. The number of sub-networks equals the number of graphs, i.e., $|G^{k}|$. We used 80\% data for training with 5-fold cross-validation. The results are averaged across 5 random seeds on the 20\% remaining test set. All models were trained in $100$ epochs with a batch size of $20$, using RMSProp optimizer~\cite{hinton2012:rmsprop} with a learning rate of $10^{-4}$. The average Mean Absolute Error (MAE), Mean Squared Error (MSE), and Root Mean Squared Error (RMSE) of 5 IMs were reported.
\paragraph{Baselines.} We compared our methods with prior works using CNN-based architecture (JOZ-CNN)~\cite{jozinovic2020:mle0}, and 2 graph-based methods, KIM-GNN~\cite{kim2021:mle1} and TSER-GCN~\cite{bloemheuvel2023:mle_main}. 
\subsection{Traffic Forecasting} 
We experimented on two well-known datasets PEMS-BAY and METR-LA~\cite{li2018:traffic_metrla_pemsbay_dcrnn} from the Bay-Area and Metropolitan LA, where the latter is known to be the harder dataset due to its more complex geography. 
\paragraph{Datasets.}The METR-LA dataset includes data from 207 sensors collected over 4 months (March 1 to June 30, 2012). The PEMS-BAY dataset includes data from 325 sensors collected over 6 months (January 1 to May 31, 2017). Traffic speed readings are aggregated into $5$-minute intervals and normalized using Z-Score. 
\paragraph{Models.} Our forecasting variants included PH-FC-AGGR$_{k}$ (defined in Section~\ref{sec:traffic-forecasting}). 
We followed the same experimental setups in the standard benchmarks such as~\cite{li2018:traffic_metrla_pemsbay_dcrnn}, including data splitting, and setting the look-back window as $T_{in}=12$. 
The models were trained for 100 epochs using the Adam optimizer~\cite{adams2017:persistence} with a learning rate of $10^{-4}$. We employed batch sizes of 64 and 32 for the METR-LA and PEMS-BAY datasets, respectively. The data is split into 70\% for training, 10\% for validation, and 20\% for testing in chronological order. We reported MAE, RMSE, and Mean Absolute Percentage Error (MAPE) on the test set.
\paragraph{Baselines.} We selected models from the leaderboard\footnote{\href{https://paperswithcode.com/sota/traffic-prediction-on-metr-la}{https://paperswithcode.com/sota/traffic-prediction-on-metr-la}} of the traffic prediction task that benchmark on both datasets, including DCRNN~\cite{li2018:traffic_metrla_pemsbay_dcrnn}, D2STGNN~\cite{shao2022:D2STGNN}, MegaRCN~\cite{jiang2023:MegaRCN}, STEP~\cite{shao2022:STEP}, RGDan~\cite{fan2024:rgdan}, and STD-MAE~\cite{gao2024:STD-MAE}. While there are many other advanced models, our goal is to demonstrate the usefulness of PH with ensembles of simple networks.
\begin{figure}[ht]
    \centering    \includegraphics[width=0.8\columnwidth]{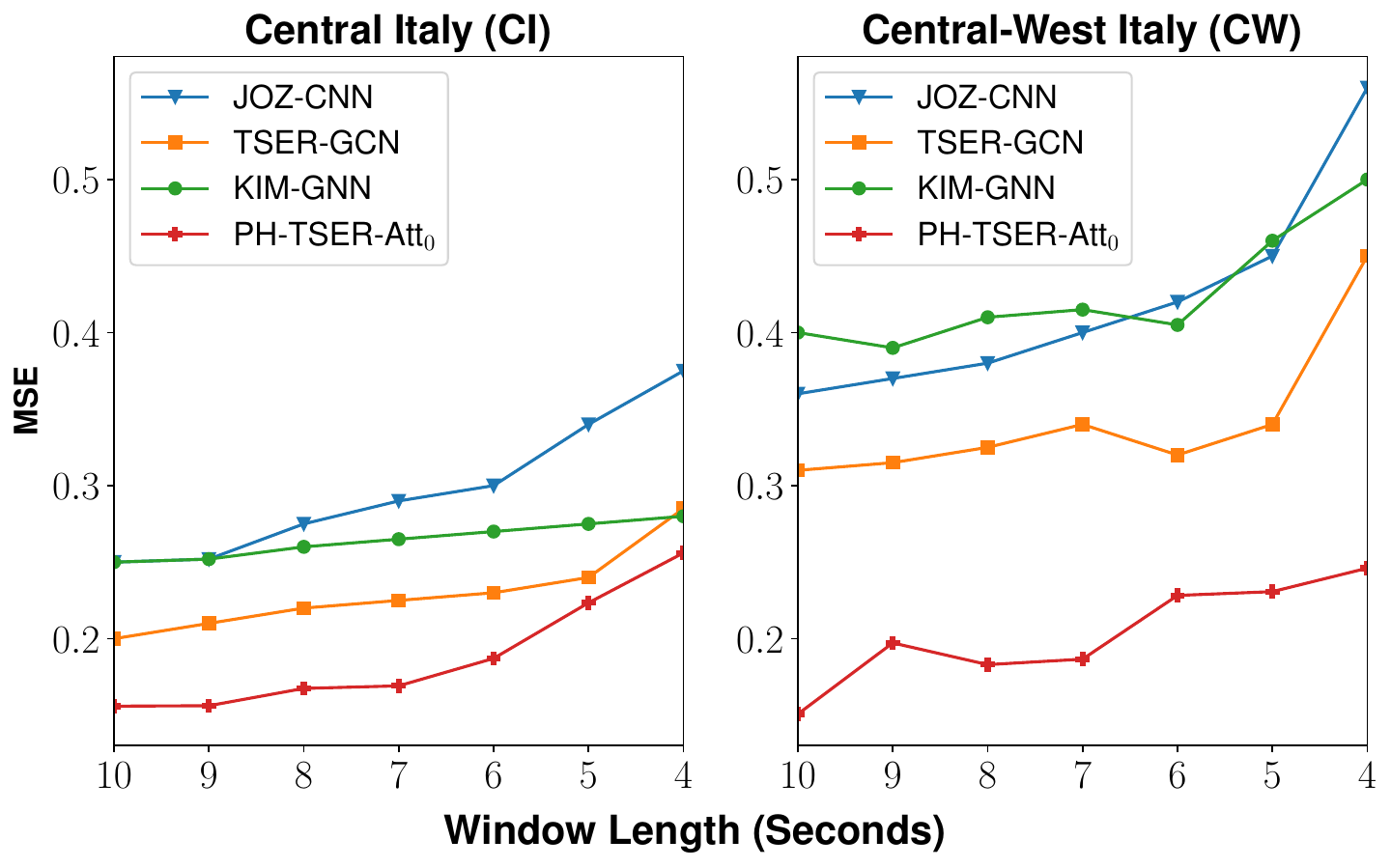}
    \caption{Results of varying window times as input.}
    \label{fig:window-reduction}
\end{figure}
\subsection{Results}
\begin{figure}[h]
    \centering
    \includegraphics[width=0.9\columnwidth]{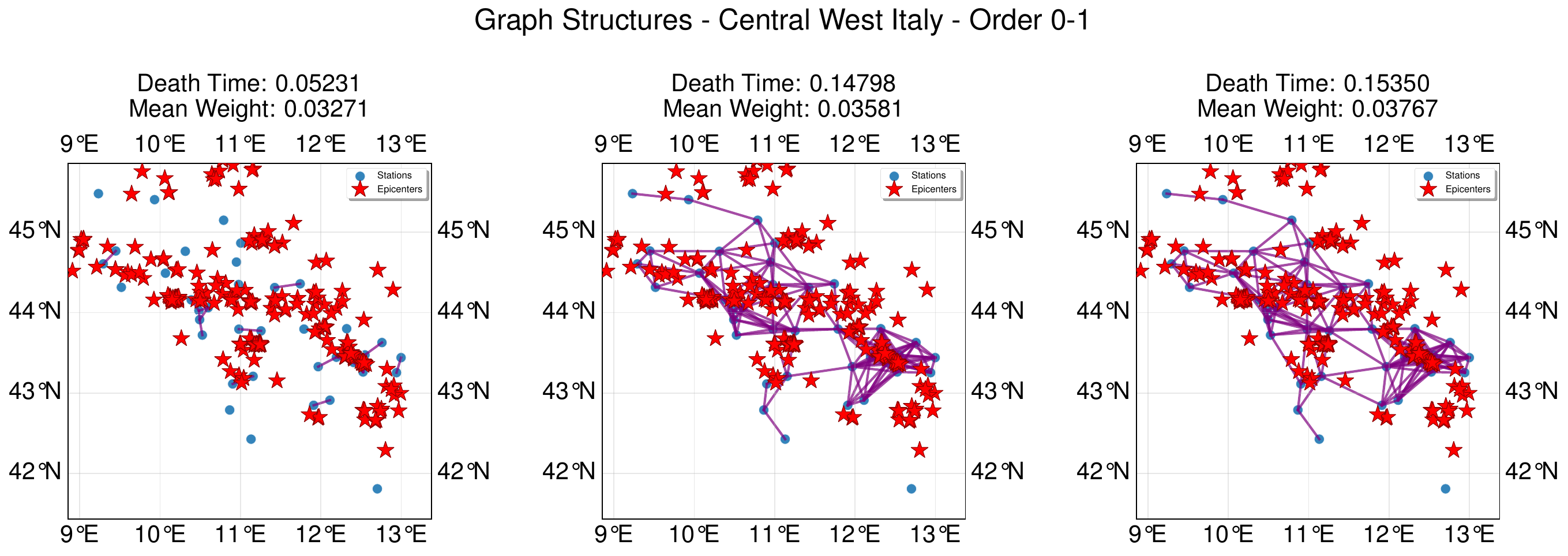}
    \caption{Graph structures on CW dataset. Three highest-weighted networks with their corresponding death times.}
    \label{fig:weighted01-CW}
\end{figure}
\paragraph{TSER on Seismic Datasets.} Table~\ref{tab:tser} reported the results of the TSER tasks on 2 earthquake datasets. All of our variants outperformed the baselines. Notably, using $G^0$ graphs (\textbf{PH-TSER-Att$_0$}) reduced the MSE of the more challenging CW network by half. This highlights the benefits of leveraging multi-scale topology for propagating sensor information. 
Following~\cite{jozinovic2020:mle0}, we conducted a window reduction sensitivity analysis, i.e. shortening the durations of the input timeseries. The minimum input time required for the model to produce meaningful predictions about the IMs was $4$ seconds. Figure~\ref{fig:window-reduction} shows that our candidate model \textbf{PH-TSER-Att}$_0$ consistently outperformed the baselines when provided with less information.
\paragraph{Traffic Forecasting.} Table~\ref{tab:traffic-forecasting} presents the performance of our models compared to the baselines on PEMS-BAY and METR-LA. Our two variants, \textbf{PH-FC-Att$_0$} and \textbf{PH-FC-Att$_{0,1}$}, demonstrate competitive results. The performance is remarkable given the simplicity of our architecture --- each temporal component is implemented as a GRU with a only a single-layer. Overall, this highlights the benefit of integrating information from multiple graph representations across multiple scales for learning tasks.
\subsection{Visualization of Graphs' Contributions}
Figure \ref{fig:weighted01-CW} demonstrates how our PH-induced graph ensemble captures seismic relationships in the Central West Italy dataset through multiple connectivity scales \footnote{Visualizations on other datasets are provided in the Appendix.}. The three network configurations, occurring at death times $0.05231$, $0.14798$, and $0.15350$ with mean weights $0.03271$, $0.03581$, and $0.03767$ respectively, reveal increasingly dense yet distinct edge patterns. While the first configuration shows selective connections between seismic stations, the subsequent configurations at higher death times develop more comprehensive connectivity patterns, particularly in regions of high seismic activity. This progression illustrates how persistent homology systematically identifies significant topological features, validating our approach of using multiple PH-derived network representations to capture complementary aspects of seismic relationships.
\section{Conclusion}
In this paper, we explored using persistent homology to generate graph-based inputs and designed ensembles of simple neural networks to fuse the extracted information for downstream tasks. Our methods proved effective in two applications: earthquake prediction and traffic speed forecasting. 

Future work could focus on developing an aggregation for the graphs before learning, which would enable using an architecture based on a single model instead of an ensemble. For example, treating graphs as distributions and solving the Barycenter problem within Optimal Transport~\cite{peyre2016:ot1,vayer2020:ot2} offers a principled approach to computing a representative graph that captures collective information from the ensemble. 


\bibliographystyle{named}
\bibliography{ijcai25}

\end{document}